\pdfoutput=1
\documentclass[11pt]{article}
\usepackage[preprint]{acl}
\usepackage{times}
\usepackage{latexsym}
\usepackage[T1]{fontenc}
\usepackage[utf8]{inputenc}
\usepackage{microtype}
\usepackage{inconsolata}
\usepackage{graphicx}
\usepackage{multirow}
\usepackage{xcolor}
\usepackage{amsmath}
\usepackage{amssymb}
\usepackage{booktabs}
\usepackage{lipsum}
\usepackage[table]{xcolor}
\definecolor{mygreen}{RGB}{1,113,0}
\usepackage{colortbl}
\definecolor{lightbluegray}{RGB}{235,245,255}
\definecolor{lightgreengray}{RGB}{235,245,235}
\usepackage{pifont}
\usepackage[utf8]{inputenc}
\usepackage{mdframed}
\usepackage{enumitem}
\usepackage[most]{tcolorbox}
\usepackage{algorithm,algpseudocode}
\usepackage{adjustbox}
\usepackage[utf8]{inputenc}
\usepackage[T1]{fontenc}   
\usepackage{hyperref}      
\usepackage{url}     
\usepackage{amsfonts}      
\usepackage{nicefrac}      
\usepackage{microtype} 
\usepackage{wrapfig}
\usepackage{arydshln}

\definecolor{customgreen}{RGB}{199,234,92}

\title{Beyond the Literal: Decomposing Pragmatic Intent \\ in Multimodal Meme Understanding}

\author{
  \textbf{Zhengyi Zhao\textsuperscript{1}},
  \textbf{Shubo Zhang\textsuperscript{1}},
  \textbf{Zezhong Wang\textsuperscript{2}},
  \textbf{Luyao Ye\textsuperscript{3}},
  \textbf{Huimin Wang\textsuperscript{4}},\\
  \textbf{Hanqi Yan\textsuperscript{5}},
  \textbf{Binyang Li\textsuperscript{6}},
  \textbf{Kam-Fai Wong\textsuperscript{1}},
  \textbf{Yulan He\textsuperscript{5}}
\\
\\
  \textsuperscript{1} The Chinese University of Hong Kong
  \textsuperscript{2} Huawei
  \textsuperscript{3} Central China Normal University\\
  \textsuperscript{4} Shenzhen University
  \textsuperscript{5} King's College London
  \textsuperscript{6} University of International Relations
\\
  {
  \texttt{zyzhao@se.cuhk.edu.hk}
  }
}

\begin{document}
\maketitle

\begin{abstract}
When asked what a meme or sarcastic post means, Large Vision Language Models (LVLMs) tend to describe what the image shows rather than what the author is trying to communicate. Standard instruction tuning entangles a post's literal content with its pragmatic meaning, letting surface-level details contaminate the final response. We reframe meme understanding as a problem of literal-pragmatic decomposition and propose \textbf{Intent Projection}, a framework that separates the two signals at the representation, output, and objective levels within a single LVLM backbone. At the representation level, an orthogonal projection module removes dominant unimodal directions from the fused image-text representation, retaining only the pragmatic residual, while a surface-real affect classifier anchors the decoder with a discrete tag that names the polarity gap. At the output level, the model externalizes a structured reasoning chain, and at the objective level a contrastive reward explicitly penalizes answers that restate the literal description. Across six multimodal benchmarks, Intent Projection consistently outperforms open-source baselines and narrows the gap to proprietary models, with the largest gains on high-divergence posts where literal collapse is most damaging.
\end{abstract}

\section{Introduction}
\label{sec:intro}

Consider a Reddit post pairing a cheerful cartoon dog with the caption ``another great Monday.'' A human reader immediately recognizes that the poster is complaining about work. The cheerfulness is the joke, not the message. Large vision language models (LVLMs), however, routinely produce literal readings of such posts by describing the dog, summarizing the caption, and missing the communicative intent entirely~\cite{maity2022multitask,wahyuni2025deep,zhao2025memereacon}. Existing works recently quantified this gap on MemeReaCon, a Reddit benchmark~\cite{zhao2025memereacon} pairing memes with their posts and community comments: even frontier systems achieve only $44.9\%$ ROUGE-L on intent generation, while surface-level classification at $83.2\%$ accuracy still falls well short of human agreement. The error analysis pinpointed the cause, which is that the models either ignore context or over-index on salient visual tokens, but left the constructive problem open: \emph{how should an LVLM be trained so that it generates from a speaker's intent rather than from what is depicted?}

We take the following view. A meme post carries two entangled signals: a \emph{literal} component, corresponding to what is shown and said, and a \emph{pragmatic} component, corresponding to why. Existing methods do address intent-related tasks~\cite{jha2024meme,zheng2025multi,bi2023you}, but they model intent as an end-to-end classification or generation target trained on the full fused representation without explicitly isolating the pragmatic signal from the literal one. The consequence is predictable: when literal features such as a smiling dog or the word ``great'' dominate the fused representation, the decoder simply redescribes what it sees, regardless of what supervision label is attached~\cite{zhao2025memereacon,lin2026goat}. Providing an intent annotation does not change which features are most accessible at decoding time. It only changes the target the model tries to reach \emph{through} those same literal features. To actually generate from the speaker's intent, the literal signal must be structurally removed from the representation before the decoder acts on it.

To this end, we propose \textbf{Intent Projection}, a framework that realizes this separation inside a single LVLM backbone along three axes. At \textbf{the representation level}, a pragmatic projection module isolates the cross-modal content of the post by removing the dominant unimodal directions from the fused image-text representation via orthogonal projection, and a surface-real affect classifier provides the decoder with a discrete conditioning tag that names the polarity gap between displayed and intended sentiment. At \textbf{the output level}, the model is trained to externalize the decomposition as an explicit reasoning chain: literal observation, intent inference, and final intent. At \textbf{the objective level}, the segments of the chain are then reinforced through a group-relative reinforcement learning stage whose reward directly penalizes literal collapse. To support training, we pair the method with a self-distillation procedure that elicits author-intent targets from the backbone itself, avoiding human annotation at scale. To support evaluation and training, we introduce a divergence metric that separates posts by how far their intended meaning departs from their literal content, which both stratifies evaluation and guides training-time sampling so that pragmatic ability is measured and targeted rather than confounded with surface understanding.

We evaluate across six benchmarks spanning meme understanding and multimodal sarcasm. Intent Projection improves over strong open-source backbones and closes a portion of the gap to frontier proprietary models, with the largest gains concentrated on high-divergence posts. Controlled ablations confirm that the reinforcement-learning stage, rather than supervised fine-tuning (SFT) alone, is responsible for the gains in the high-divergence regime, and that the contrastive reward component is what isolates pragmatic from literal content. In summary, our contributions are as follows:
\begin{itemize}
    \item We reframe contextual meme understanding as a problem of \emph{literal-pragmatic decomposition} over the generated chain, identifying the contamination of intent by literal content as the mechanism behind the failure mode documented by prior work.
    \item We propose \textbf{Intent Projection}, which explicitly models the literal-pragmatic separation at the representation level by projecting out unimodal literal directions from the fused representation, giving the decoder direct access to a pragmatic residual.
    \item Across six benchmarks spanning memes and multimodal sarcasm, Intent Projection consistently outperforms strong open-source baselines and closes a substantial portion of the gap to frontier proprietary systems, with gains concentrated in the high-divergence regime.
\end{itemize}

\section{Related Work}
\label{sec:related_works}

\paragraph{Meme understanding and multimodal sarcasm.}
Recent work~\cite{nguyen2024computational,xu2024generating,li2026kid,de2026m} highlights that LVLMs struggle when a post's intent departs from its literal content. MemeReaCon~\cite{zhao2025memereacon} documents a large gap between surface classification and intent generation even for frontier systems. Most methods~\cite{zheng2025multi,bi2023you,jha2024meme,niu2026behind,li2026few} rely on early cross-attention fusion, entangling literal and pragmatic signals with no explicit handle on the surface-real gap. Sentiment-conditioned generation partially addresses this but uses a single polarity label, conflating sarcasm with sincere positivity; our two-axis surface$\times$real formulation captures the \emph{reversal} between displayed and intended affect.

\paragraph{Disentangled representations and RL for LVLM reasoning.}
Orthogonal projection~\cite{lin2025matryoshkakv} and subspace methods~\cite{liu2025incomplete,cai2025learning,sun2025multimodal} separate shared from modality-specific content but operate purely at the representation level with discriminative heads. On the optimization side, GRPO~\cite{mroueh2025revisiting} elicits chain-of-thought reasoning without a value network, and recent extensions~\cite{lee2025sample,yuan2026more,adams2026rethinking,nguyen2025aligning,wang2026top} apply it to multimodal settings with task-correctness rewards. Our contribution is orthogonal: we introduce a contrastive, literal-aware reward that penalizes similarity between the model's final answer and its own literal segment, directly targeting literal collapse rather than task accuracy alone.

\section{Methodology}
\label{sec:method}

\begin{figure*}[!t]
    \centering
    \includegraphics[width=\textwidth]{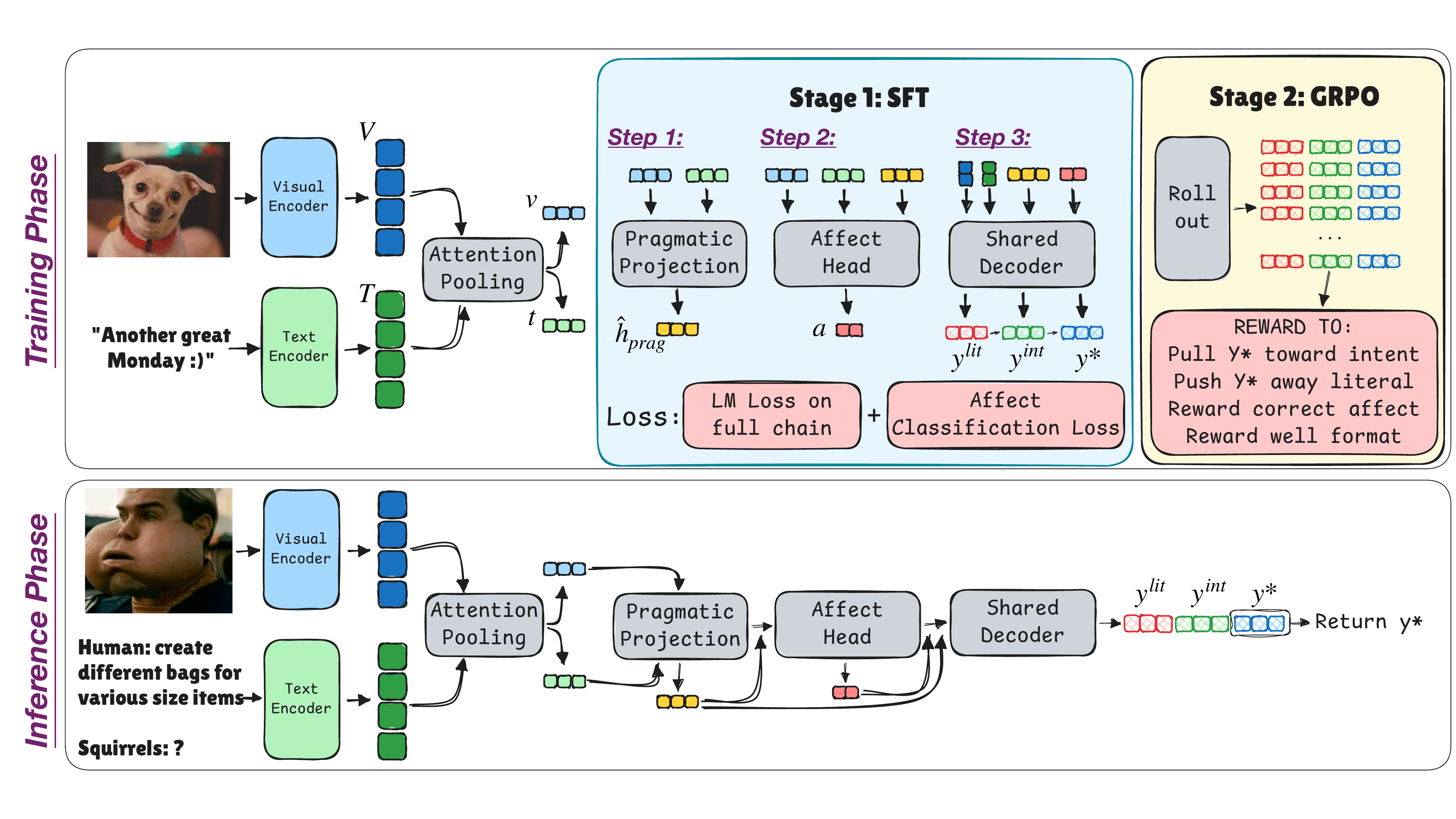} 
    \caption{Overview of the Intent Projection. The model uses an LVLM backbone augmented with a Pragmatic Projection module and a Surface-Real Affect Head. Training involves a two-stage process: Stage 1 utilizes SFT for format teaching, while Stage 2 applies GRPO with a contrastive, literal-aware reward to prevent literal collapse. During the inference phase, the model only requires the meme post $x = (x_v, x_t)$ to autoregressively generate the full structured chain, ultimately extracting the verbalized communicative intent $y^\star$.}
    \label{fig:overview}
\end{figure*}

Given a meme post $x=(x_v, x_t)$ consisting of an image $x_v$ and its accompanying text $x_t$, our goal is to produce an utterance $y^\star$ that verbalizes the \emph{author's communicative intent} rather than a literal description of $x$. Figure~\ref{fig:overview} gives an overview. 

\subsection{Problem Formulation}
\label{subsec:formulation}

We treat a meme post as carrying two entangled signals. Let $h \in \mathbb{R}^{d}$ denote a joint representation of $x$ produced by an LVLM. We posit the latent decomposition
\begin{equation}
  h = h_{\text{lit}} + h_{\text{prag}} + \epsilon,
  \label{eq:decomp}
\end{equation}
where $h_{\text{lit}}$ captures what is \emph{depicted and said}, $h_{\text{prag}}$ captures \emph{why} the post was made, and $\epsilon$ absorbs residual variation. Standard instruction tuning learns $p_\theta(y|h)$ directly. Because the vision and language encoders are pretrained to faithfully represent depicted content, $h_{\text{lit}}$ is the most readily accessible component of $h$ at decoding time, making the decoder prone to literal generation even when the training target is pragmatic. Intent Projection instead estimates $\widehat{h}_{\text{prag}}$ and conditions the intent-bearing segment of the output on $\widehat{h}_{\text{prag}}$ rather than on $h$. We do not claim $\widehat{h}_{\text{prag}}$ recovers $h_{\text{prag}}$ pointwise but require that it carries enough pragmatic information to redirect generation away from $h_{\text{lit}}$.

\subsection{Pragmatic Projection}
\label{subsec:projection}

We build on a decoder-only LVLM exposing a visual encoder $f_v$, a text embedder $f_t$, and a shared transformer decoder $g_\theta$. Given $x=(x_v, x_t)$, the encoders yield token sequences $V = f_v(x_v) \in \mathbb{R}^{n_v \times d}$ and $T = f_t(x_t) \in \mathbb{R}^{n_t \times d}$, from which we obtain pooled summaries $v = \mathrm{Pool}(V)$ and $t = \mathrm{Pool}(T)$ via attention pooling. These summaries are used \emph{only} by the projection module below. The full token sequences $V, T$ are still passed to $g_\theta$, so the decoder retains its standard grounding pathway.

Consider again the smiling cartoon dog paired with ``another great Monday.'' The sarcasm arises not from the smile alone nor from the word ``Monday'' alone, but from the \emph{tension} between a positive visual cue and a negative situational referent. Each modality in isolation is literal. Pragmatic meaning emerges only from how the two modalities interact. More generally, what each modality conveys on its own, the appearance of the dog or the dictionary meaning of the caption, constitutes the literal component $h_{\text{lit}}$, while the meaning that arises only when image and text are read \emph{together} constitutes the pragmatic component $h_{\text{prag}}$. We therefore define $\widehat{h}_{\text{prag}}$ as the part of the fused representation that cannot be explained by either modality independently. We compute $\widehat{h}_{\text{prag}}$ in two steps. First, we project $v$ and $t$ into a shared interaction space via learnable linear maps $W_v, W_t \in \mathbb{R}^{d \times d}$ that extract modality-specific features relevant to cross-modal interaction, and fuse the results into a single vector that captures what the two modalities convey together:
\begin{equation}
  c = \sigma\!\left(W_s\left[W_v v \,\Vert\, W_t t\right]\right),
\end{equation}
where $W_s \in \mathbb{R}^{d \times 2d}$ maps the concatenation back to $d$ dimensions. Second, we subtract from $c$ the directions along which each modality summarizes itself, keeping only what remains. Let $u_v = W_v v / \|W_v v\|_2$ and $u_t = W_t t / \|W_t t\|_2$ be the unit-length summaries of the two modalities, and let $U = [u_v \;\; u_t]$ span the resulting \emph{literal subspace} $\mathcal{L}$. We take $\widehat{h}_{\text{prag}}$ to be the component of $c$ orthogonal to this subspace:
\begin{equation}
  \widehat{h}_{\text{prag}} = \bigl(I - P_{\mathcal{L}}\bigr)\, c, \qquad P_{\mathcal{L}} = U(U^\top U)^{-1} U^\top.
  \label{eq:orth}
\end{equation}
We feed $\widehat{h}_{\text{prag}}$ to the decoder as a single soft token prepended to the input, with its own learned type embedding. We put more discussions in Appendix~\ref{apd:method}.

\subsection{Surface-Real Affect Conditioning}
\label{subsec:affect}

The pragmatic vector $\widehat{h}_{\text{prag}}$ already encodes the cross-modal signal that drives intent, but a single continuous prefix token must compete with a growing sequence of generated tokens for the decoder's attention budget. In practice, we observe that relying solely on this prefix leads to degraded intent generation on longer outputs. A discrete symbolic tag, by contrast, is re-encoded at each position and thus provides a persistent anchor throughout generation. We therefore augment $\widehat{h}_{\text{prag}}$ with a discrete affect tag that names the surface-real relationship explicitly. The tag is a \emph{supplement}, not a substitute: $\widehat{h}_{\text{prag}}$ provides fine-grained continuous signal, the tag provides a coarse symbolic anchor. We define affect over the four-way Cartesian product
\begin{equation}
  a \in \mathcal{A} = \{\text{pos},\text{neg}\}_{\text{surface}} \times \{\text{pos},\text{neg}\}_{\text{real}},
\end{equation}
yielding \textsc{Pos-Pos} (sincere positive), \textsc{Neg-Neg} (sincere negative), \textsc{Pos-Neg} (sarcasm), and \textsc{Neg-Pos} (dark joy). The two-axis structure is essential: a single-axis sentiment head collapses sarcasm and sincere positivity into the same label and therefore cannot represent the gap that defines the failure mode we target. The diagonal classes correspond to literal posts; the off-diagonal classes concentrate in the high-divergence regime.

Then a two-layer MLP head $\phi_a$ produces
\begin{equation}
  p_a = \mathrm{softmax}\!\bigl(\phi_a([v\,;\, t\,;\, \widehat{h}_{\text{prag}}])\bigr),
\end{equation}
where $v, t$ are the pooled modality summaries defined in \S\ref{subsec:projection}. The argmax label is linearized as a short tag (e.g.\ \texttt{[affect: surface-pos, real-neg]}) and prepended to the decoder prompt together with the soft token for $\widehat{h}_{\text{prag}}$. The decoder thus receives two complementary handles on the post: a continuous vector that captures fine-grained cross-modal content, and a discrete symbol that names the surface-real relationship. The two provide complementary granularities--continuous and symbolic--so the decoder benefits from both throughout generation.

\subsection{Training and Inference}
\label{subsec:training}

We train the backbone, projection parameters, and affect head end-to-end in two stages following standard SFT and GRPO recipes. What is task-specific is that the model emits a three-segment chain delimited by special tokens: a \emph{literal} segment $y^{\text{lit}}$ describing what the image shows and what the text says, an \emph{intent reasoning} segment $y^{\text{int}}$ resolving why the two were paired, and a \emph{final intent} segment $y^\star$. In Stage~1 we optimize the following loss jointly over the backbone, $\{W_v, W_t, W_s\}$, and $\phi_a$:
\begin{equation}
\mathcal{L}_{\text{SFT}} = \mathcal{L}_{\text{LM}}(y^{\text{lit}}, y^{\text{int}}, y^\star) + \gamma\, \mathcal{L}_{\text{CE}}(p_a, a^\star).
\end{equation}
Targets $(y^{\text{lit}}, y^{\text{int}}, y^\star, a^\star)$ are obtained via self-distillation: we prompt the backbone with the image, post title, and top-voted community comments as auxiliary context, asking it to produce the three-segment chain. The resulting $y^\star$ serves as the silver-standard reference intent $r^\star$ for Stage~2. Crucially, community comments are used \emph{only} during this target-elicitation step and are never visible to the model at training or inference time.

SFT teaches the chain format but does not prevent literal collapse: the model emits the chain while still letting $y^{\text{lit}}$ dominate $y^\star$. Stage~2 addresses this with GRPO under a contrastive, literal-aware reward
\begin{equation}
\begin{aligned}
  R(y^{(i)}) \;=\;& \mathrm{sim}(y^{\star,(i)}, r^\star) \;-\; \lambda_1\, \mathrm{sim}(y^{\star,(i)}, y^{\text{lit},(i)}) \\
  &+\; \lambda_2\, \mathbf{1}[a^{(i)} = a^\star] \;+\; \lambda_3\, R_{\text{fmt}}(y^{(i)}),
\end{aligned}
\label{eq:reward}
\end{equation}
where $\mathrm{sim}(\cdot,\cdot)$ is cosine similarity from a frozen sentence encoder (bounded in $[-1, 1]$), $\mathbf{1}[\cdot] \in \{0,1\}$, and $R_{\text{fmt}} \in \{0,1\}$ indicates whether the output contains all three chain segments with correct delimiters. All terms are bounded, so the $\lambda$ weights control relative importance directly. The first term pulls $y^\star$ toward the reference intent; the second pushes $y^\star$ \emph{away} from the model's own literal rendering, making the two segments compete rather than collapse; the remaining terms reward affect consistency and chain format. Per-token advantage weights are zeroed for all positions in $y^{\text{lit}}$ (equivalent to stop-gradient on the literal segment), so only $y^{\text{int}}$ and $y^\star$ receive GRPO updates while $y^{\text{lit}}$ is held at its SFT distribution, anchoring the reference point of the contrast. To measure and target pragmatic ability rather than surface understanding, we define a literal-intent divergence score:
\begin{equation}
\Delta(x) = 1 - \mathrm{sim}(\ell(x), r^\star(x)),
\end{equation}
where $\ell(x)$ is a literal caption obtained by prompting the backbone without community context. 
At inference, only $x = (x_v, x_t)$ is required: the encoders produce $V, T$ and pooled summaries $v, t$; the projection module computes $\widehat{h}_{\text{prag}}$ via Eq.~\ref{eq:orth}; the affect head emits a tag from $p_a$; and the decoder, conditioned on $V, T$, the soft token for $\widehat{h}_{\text{prag}}$, and the linearized affect tag, generates the chain autoregressively, returning the \texttt{<ans>} segment as the final intent.

\section{Experiments}
\label{sec:experiments}

We organize experiments around four questions. \textbf{Q1}: does Intent Projection improve pragmatic intent generation on the benchmark? \textbf{Q2}: does the surface-real affect head produce a genuinely useful pragmatic signal? \textbf{Q3}: does the contrastive, literal-aware reward concentrate gains on high-divergence posts? \textbf{Q4}: are the gains robust to backbone, scale, silver-target provenance, and reward weights?

\subsection{Setup}
\label{subsec:setup}

\paragraph{Datasets.} We evaluate on six benchmarks grouped into three categories. \textbf{Contextual meme understanding}: MemeReaCon~\cite{zhao2025memereacon} provides four tasks over Reddit posts: Post Connections Generation (PC-G), Post Intent Generation (PI-G), Context-Meme Interplay Classification (CMI-C), and Comment Stance and Affective Consistence Classification (CSAC-C). \textbf{Meme explanation}: MemeCap~\cite{hwang2023memecap} shows context understanding and MET-Meme~\cite{xu2022met} is for intent classification. \textbf{Multimodal sarcasm}: MMSD2.0~\cite{qin2023mmsd2} for binary sarcasm classification, MuSE~\cite{jing2023multi} for sarcasm explanation, and GOAT-Bench~\cite{lin2026goat} for contextual abusive-meme understanding. Preprocessing and prompt templates are in Appendix~\ref{app:data}.

\paragraph{Backbones and metrics.} We apply Intent Projection to three open-source LVLMs: LLaVA-OneVision-7B~\cite{li2024llava}, Qwen3-VL-8B~\cite{bai2025qwen3}, and InternVL3.5-8B~\cite{wang2025internvl3} and frontier proprietary models. For generation tasks we report BERTScore-F1 (B-S) and ROUGE-L (R-L); for classification, Accuracy and macro-F1. Full details are in Appendix~\ref{app:std}.

\paragraph{Training variants.} For each backbone we report: (i) \textbf{Base} zero-shot; (ii) \textbf{CoT} prompting; (iii) \textbf{SFT} without chain structure; (iv) \textbf{SFT\,+\,Chain}; (v) \textbf{SFT+Chain+GRPO$_\text{flat}$} with intent-match reward only; (vi) \textbf{Ours (full)}.

\subsection{Main Results}
\label{subsec:main_gen}

\begin{table*}[!t]
\centering
\small
\adjustbox{max width=\linewidth}{
\begin{tabular}{lcccccccccccc}
\toprule
\multirow{2}{*}{\textbf{Model}} & \multicolumn{2}{c}{\textbf{MRC PC-G}} & \multicolumn{2}{c}{\textbf{MRC PI-G}} & \multicolumn{2}{c}{\textbf{MemeCap}} & \multicolumn{2}{c}{\textbf{MET-Meme}} & \multicolumn{2}{c}{\textbf{MuSE}} & \multicolumn{2}{c}{\textbf{GOAT-G}} \\
 & B-S & R-L & B-S & R-L & B-S & R-L & B-S & R-L & B-S & R-L & B-S & R-L \\
\midrule
\multicolumn{13}{l}{\textit{Frontier proprietary (zero-shot, reference only)}} \\
GPT-5.4-mini & 57.8 & 48.5 & 34.9 & 28.4 & 51.7 & 32.1 & 49.8 & 30.6 & 54.2 & 36.5 & 52.4 & 33.8 \\
Claude-Sonnet-4.6 & 64.5 & 57.1 & 47.6 & 40.3 & 58.3 & 40.4 & 55.9 & 38.2 & 60.7 & 43.9 & 58.0 & 41.2 \\
Gemini-3-Flash & 66.9 & 60.4 & 52.3 & 44.9 & 60.8 & 43.0 & 58.4 & 40.7 & 62.5 & 46.1 & 60.1 & 43.7 \\
\midrule
{\textit{LLaVA-OneVision-7B}} & 20.2 & 22.5 & 12.7 & 10.9 & 38.4 & 18.2 & 36.7 & 17.9 & 41.3 & 22.0 & 37.9 & 19.1 \\
\quad + CoT & 26.1 & 27.7 & 15.8 & 13.5 & 40.1 & 20.3 & 38.2 & 19.6 & 43.1 & 24.1 & 39.4 & 20.8 \\
\quad + SFT & 33.8 & 34.4 & 19.7 & 16.8 & 42.9 & 23.2 & 41.0 & 22.1 & 46.5 & 27.4 & 42.6 & 24.0 \\
\quad + SFT\,+\,Chain & 36.2 & 36.7 & 21.8 & 18.4 & 44.1 & 24.3 & 42.4 & 23.0 & 47.9 & 28.6 & 43.9 & 25.1 \\
\quad + SFT+Chain+GRPO$_\text{flat}$ & 40.1 & 39.8 & 23.6 & 20.1 & 45.8 & 25.7 & 44.0 & 24.3 & 49.5 & 30.0 & 45.6 & 26.6 \\
\quad \textbf{+ Ours (full)} & \textbf{43.6} & \textbf{43.1} & \textbf{25.9} & \textbf{22.1} & \textbf{47.8} & \textbf{27.5} & \textbf{45.7} & \textbf{25.8} & \textbf{51.6} & \textbf{32.0} & \textbf{47.4} & \textbf{28.3} \\
\midrule
{\textit{Qwen3-VL-8B}} & 33.7 & 31.3 & 19.8 & 16.0 & 43.6 & 22.7 & 41.2 & 21.5 & 46.8 & 26.2 & 42.1 & 22.5 \\
\quad + CoT & 38.1 & 35.4 & 23.6 & 19.1 & 45.2 & 24.6 & 43.0 & 23.4 & 48.5 & 28.3 & 44.0 & 24.6 \\
\quad + SFT & 47.3 & 44.9 & 29.1 & 24.2 & 48.6 & 27.5 & 46.3 & 25.8 & 52.1 & 32.0 & 47.4 & 27.8 \\
\quad + SFT\,+\,Chain & 50.1 & 47.6 & 31.9 & 26.7 & 49.8 & 28.7 & 47.6 & 26.9 & 53.4 & 33.3 & 48.6 & 28.9 \\
\quad + SFT+Chain+GRPO$_\text{flat}$ & 53.5 & 50.9 & 34.2 & 29.5 & 51.5 & 30.2 & 49.2 & 28.3 & 55.1 & 34.8 & 50.2 & 30.4 \\
\quad \textbf{+ Ours (full)} & \textbf{57.4} & \textbf{54.3} & \textbf{37.2} & \textbf{32.8} & \textbf{53.7} & \textbf{32.4} & \textbf{51.0} & \textbf{30.1} & \textbf{57.2} & \textbf{37.1} & \textbf{52.0} & \textbf{32.3} \\
\midrule
{\textit{InternVL3-8B}} & 37.2 & 38.9 & 25.6 & 20.4 & 45.1 & 24.3 & 42.8 & 23.0 & 48.2 & 27.6 & 43.9 & 24.1 \\
\quad + CoT & 41.8 & 42.5 & 28.9 & 23.2 & 46.7 & 26.1 & 44.3 & 24.7 & 49.9 & 29.4 & 45.5 & 26.0 \\
\quad + SFT & 50.3 & 48.6 & 33.8 & 27.6 & 49.5 & 28.4 & 47.0 & 26.6 & 53.0 & 32.9 & 48.3 & 29.2 \\
\quad + SFT\,+\,Chain & 52.9 & 51.0 & 36.4 & 29.9 & 50.7 & 29.5 & 48.2 & 27.7 & 54.3 & 34.1 & 49.6 & 30.3 \\
\quad + SFT+Chain+GRPO$_\text{flat}$ & 56.2 & 54.1 & 38.9 & 32.6 & 52.4 & 31.0 & 49.8 & 29.1 & 56.0 & 35.7 & 51.2 & 31.9 \\
\quad \textbf{+ Ours (full)} & \textbf{59.8} & \textbf{57.2} & \textbf{41.3} & \textbf{35.6} & \textbf{54.5} & \textbf{33.1} & \textbf{51.7} & \textbf{30.8} & \textbf{58.0} & \textbf{37.8} & \textbf{53.1} & \textbf{33.8} \\
\bottomrule
\end{tabular}}
\caption{Generation results across six tasks. B-S indicates BERTScore-F1, and R-L denotes ROUGE-L (\%). Proprietary models are zero-shot reference points. Best open-source numbers are in \textbf{bold}.}
\label{tab:main_gen}
\end{table*}

\begin{table*}[!t]
\centering
\small
\adjustbox{max width=\linewidth}{
\begin{tabular}{lcccccccc}
\toprule
\multirow{2}{*}{\textbf{Model}} & \multicolumn{2}{c}{\textbf{MRC CMI-C}} & \multicolumn{2}{c}{\textbf{MRC CSAC-C}} & \multicolumn{2}{c}{\textbf{MMSD2.0}} & \multicolumn{2}{c}{\textbf{GOAT-C}} \\
 & Acc & F1 & Acc & F1 & Acc & F1 & Acc & F1 \\
\midrule
GPT-5.4-mini & 72.5 & 72.0 & 58.8 & 46.5 & 82.4 & 81.8 & 74.6 & 73.1 \\
Claude-Sonnet-4.6 & 81.0 & 80.6 & 68.4 & 55.9 & 85.2 & 84.6 & 79.3 & 78.2 \\
Gemini-3-Flash & 83.2 & 82.9 & 71.3 & 59.4 & 86.9 & 86.2 & 81.5 & 80.4 \\
\midrule
LLaVA-OneVision-7B & 56.3 & 55.8 & 38.9 & 29.1 & 71.3 & 70.5 & 65.4 & 64.2 \\
\quad + SFT\,+\,Chain & 66.8 & 66.2 & 49.0 & 37.8 & 78.4 & 77.6 & 72.3 & 71.1 \\
\quad \textbf{+ Ours} & \textbf{70.5} & \textbf{70.0} & \textbf{53.7} & \textbf{41.6} & \textbf{80.9} & \textbf{80.1} & \textbf{74.8} & \textbf{73.7} \\
Qwen3-VL-8B & 61.4 & 61.0 & 46.2 & 35.8 & 74.8 & 74.0 & 68.7 & 67.5 \\
\quad + SFT\,+\,Chain & 72.6 & 72.1 & 57.3 & 44.2 & 81.2 & 80.4 & 75.4 & 74.3 \\
\quad \textbf{+ Ours} & \textbf{76.4} & \textbf{75.9} & \textbf{61.8} & \textbf{47.9} & \textbf{83.7} & \textbf{82.9} & \textbf{78.1} & \textbf{77.0} \\
InternVL3-8B & 64.7 & 64.2 & 49.5 & 38.4 & 76.1 & 75.3 & 70.0 & 68.9 \\
\quad + SFT\,+\,Chain & 75.9 & 75.3 & 60.4 & 46.9 & 82.5 & 81.7 & 76.8 & 75.6 \\
\quad \textbf{+ Ours} & \textbf{79.6} & \textbf{79.1} & \textbf{65.0} & \textbf{50.8} & \textbf{84.8} & \textbf{84.0} & \textbf{79.4} & \textbf{78.3} \\
\bottomrule
\end{tabular}}
\caption{Classification results (Accuracy and Macro-F1, \%). Best open-source numbers per column in \textbf{bold}.}
\label{tab:main_cls}
\end{table*}

\paragraph{Generation Tasks.} Table~\ref{tab:main_gen} reports generation results across MemeReaCon PC-G and PI-G along with MemeCap, MET-Meme, MuSE, and GOAT-G. Intent Projection yields consistent gains across all three backbones and all six tasks, with the full pipeline exceeding SFT+Chain+GRPO$_\text{flat}$ by 2-4\% ROUGE-L. This gap isolates the contrastive, literal-aware reward as the component responsible for the final performance increment. The intermediate variants reveal a clear progression: CoT and SFT each contribute partial gains, but neither alone closes the gap to the full method, and SFT+Chain still trails Ours by 5-7\% ROUGE-L on the harder generation tasks. Notably, on PI-G our 8B-scale Qwen3-VL with Intent Projection reaches 32.8\% ROUGE-L, narrowing the zero-shot gap to Gemini-3-Flash by roughly 60\% despite being more than an order of magnitude smaller.

\paragraph{Classification Tasks.} Table~\ref{tab:main_cls} reports classification results. Intent Projection improves all three backbones, with the largest gains on CSAC-C. This is expected because CSAC-C requires recognizing whether a comment's stance is consistent with the post's \emph{intended} affect, exactly the surface-real distinction our affect head models. On MMSD2.0, our method raises Qwen3-VL-8B from $74.8\%$ to $83.7\%$ accuracy. Training details for all backbones, including reward convergence and literal-intent separation over the course of GRPO, are detailed in Appendix~\ref{apd:full_training}.

%
%

\begin{figure*}[!t]
\centering
\includegraphics[width=\linewidth]{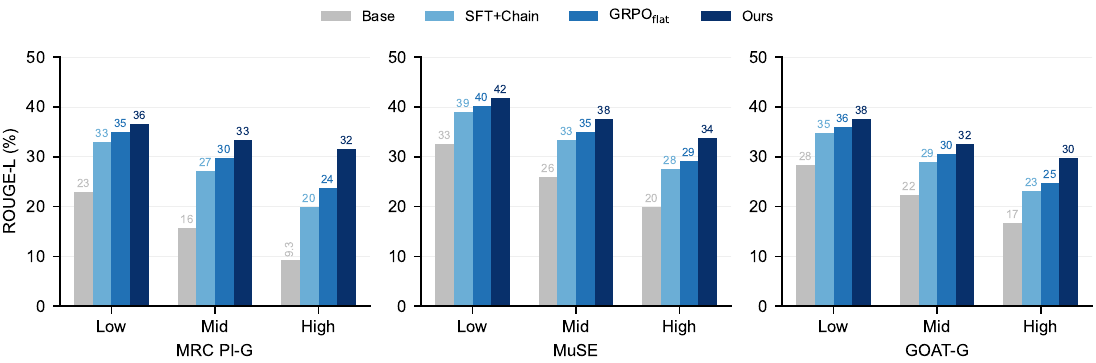}
\caption{Divergence-stratified ROUGE-L on Qwen3-VL-8B backbone. Posts split into Low/Mid/High categories by literal-intent divergence. Numbers atop bars show absolute scores. Our method maintains performance on High-$\Delta$ posts where baselines collapse.}
\label{fig:divergence}
\end{figure*}

\subsection{Analysis on Affect Head}
\label{subsec:affect_probe}

Table~\ref{tab:affect} isolates the affect head's contribution on the MemeReaCon inconsistent-affect subset (posts where surface sentiment contradicts intended sentiment) using Qwen3-VL-8B. We draw two conclusions. First, removing the affect head drops four-way affect accuracy (Probe~A) by $14.9$\% and downstream PI-G ROUGE-L (Probe~B) by $4.4$\%, confirming that the head provides a signal the decoder actively relies on. Second, collapsing the two-axis formulation into a single binary sentiment axis causes a comparable drop. This demonstrates that sarcasm is not a matter of final sentiment polarity. What matters is the \emph{reversal relationship} between displayed and intended affect. A single-axis head conflates sarcasm with sincere positivity because both map to the same output label, erasing precisely the distinction the decoder needs.

\begin{table}[!t]
\centering
\small
\adjustbox{max width=\linewidth}{
\begin{tabular}{lcc}
\toprule
\textbf{Variant} & \textbf{Probe A (Acc)} & \textbf{Probe B (R-L)} \\
\midrule
Full method & 64.2 & 31.6 \\
\; w/o affect head & 49.3 & 27.2 \\
\; w/o reward term $\lambda_2$ only & 63.8 & 29.0 \\
\; 2-axis $\to$ 1-axis binary & 51.7 & 27.6 \\
\; w/o head \& $\lambda_2$ & 49.0 & 25.9 \\
\bottomrule
\end{tabular}}
\caption{Affect-head probes on MemeReaCon inconsistent-affect subset. Probe A indicates four-way affect accuracy; Probe B shows PI-G ROUGE-L on the same subset.}
\label{tab:affect}
\end{table}

\begin{table*}[!t]
\centering
\adjustbox{max width=\linewidth}{
\begin{tabular}{lcccc}
\toprule
\textbf{Variant} & \textbf{PI-G all} & \textbf{PI-G high-$\Delta$} & \textbf{MMSD2.0} & \textbf{CSAC-C F1} \\
\midrule
Full method & 32.8 & 31.6 & 83.7 & 47.9 \\
\midrule
\; w/o divergence sampling & 32.0 {\color{gray}\scriptsize($-$0.8)} & 29.4 {\color{gray}\scriptsize($-$2.2)} & 83.5 & 47.6 \\
\; w/o orthogonal projection & 31.1 {\color{gray}\scriptsize($-$1.7)} & 28.5 {\color{gray}\scriptsize($-$3.1)} & 82.1 & 46.4 \\
\; w/o affect head & 30.9 {\color{gray}\scriptsize($-$1.9)} & 28.0 {\color{gray}\scriptsize($-$3.6)} & 82.4 & 43.1 \\
\; w/o GRPO stage & 26.7 {\color{gray}\scriptsize($-$6.1)} & 19.9 {\color{gray}\scriptsize($-$11.7)} & 81.2 & 44.2 \\
\; w/o contrastive $\lambda_1$ & 29.5 {\color{gray}\scriptsize($-$3.3)} & 23.8 {\color{gray}\scriptsize($-$7.8)} & 82.6 & 47.1 \\
\; w/o chain structure & 24.2 {\color{gray}\scriptsize($-$8.6)} & 17.6 {\color{gray}\scriptsize($-$14.0)} & 80.9 & 42.8 \\
\midrule
\; w/o gradient masking & 30.2 {\color{gray}\scriptsize($-$2.6)} & 25.3 {\color{gray}\scriptsize($-$6.3)} & 82.9 & 46.5 \\
\; w/o dropout $p_\text{drop}{=}0$ & 32.1 {\color{gray}\scriptsize($-$0.7)} & 30.4 {\color{gray}\scriptsize($-$1.2)} & 83.1 & 47.3 \\
\; affect tag withheld at inference & 31.4 {\color{gray}\scriptsize($-$1.4)} & 29.2 {\color{gray}\scriptsize($-$2.4)} & 82.7 & 45.0 \\
\bottomrule
\end{tabular}}
\caption{Component ablation on Qwen3-VL-8B. Gray numbers denote drop from full method.}
\label{tab:ablation}
\end{table*}

\subsection{Results on Posts Where Intent Differs from the Image}
\label{subsec:divergence}

The central hypothesis of this work is that literal collapse is most damaging on posts whose intended meaning sharply diverges from their literal content. We test this by computing $\Delta(x) = 1 - \mathrm{sim}(\ell(x), r^\star(x))$ for each test post and splitting into Low/Mid/High terciles. Figure~\ref{fig:divergence} confirms the hypothesis: base models degrade sharply from Low to High, whereas our method produces a nearly flat profile with the gap reduced below 5\%. This demonstrates that Intent Projection does not merely improve average performance, it specifically resolves the failure mode it targets, maintaining generation quality precisely where baselines collapse into literal redescription. Each intermediate stage partially flattens the curve, but only the full method achieves consistent High-$\Delta$ performance. Appendix~\ref{app:divergence} extends the analysis to additional backbones.

\begin{figure}[!t]
\centering
\includegraphics[width=\linewidth]{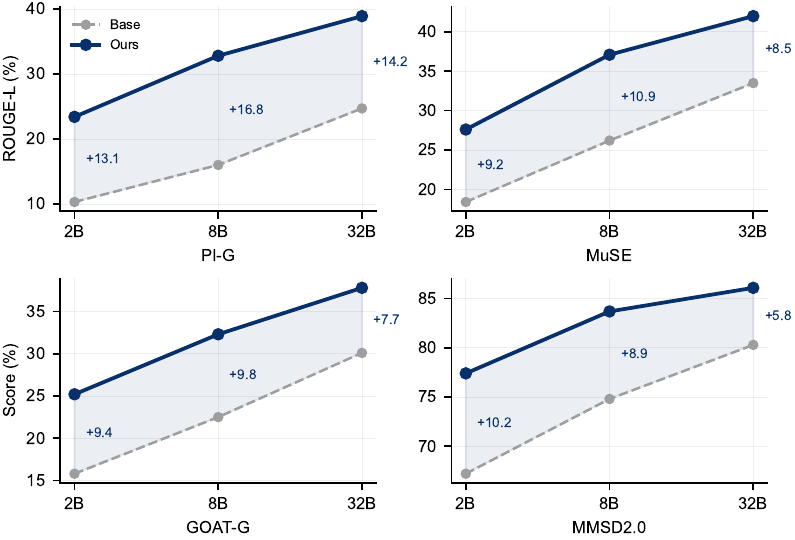}
\caption{Scaling behavior (Qwen3-VL 2B/8B/32B). Solid line denotes Ours, and dashed one shows Base. The shaded area shows performance gain.}
\label{fig:scale}
\end{figure}

\begin{figure*}[!t]
\centering
\includegraphics[width=\linewidth]{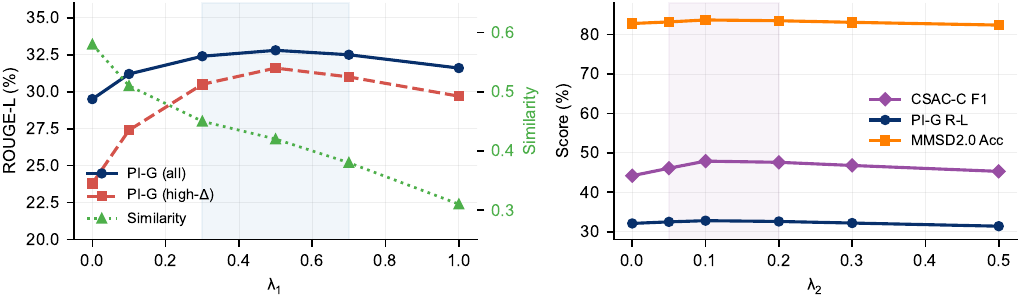}
\caption{Reward weight sensitivity. Left: $\lambda_1$ sweep with PI-G ROUGE-L (left axis) and literal-intent cosine similarity (right axis). Right: $\lambda_2$ sweep across three metrics.}
\label{fig:lambda}
\end{figure*}

\subsection{Ablation Study}
\label{subsec:ablation}

Table~\ref{tab:ablation} shows the ablation study. We highlight three conclusions. First, the GRPO stage is indispensable. Removing it causes the largest single drop, confirming that SFT alone teaches the chain format but does not prevent literal collapse. Second, the contrastive term $\lambda_1$ shows a disproportionate impact on High-$\Delta$ posts relative to overall, demonstrating it targets precisely the regime where literal and pragmatic signals diverge most. Third, removing the affect head entirely ($-1.9$) hurts more than withholding the tag only at inference ($-1.4$), yet the inference-only removal still degrades performance noticeably, confirming the decoder learns to actively condition on the tag during generation. Gradient masking on $y^{\text{lit}}$ anchors the contrastive reward by holding the literal segment stable as a reference point. Without it, both segments drift together and the contrast loses meaning. Further chain-segment diagnostics are detailed in Appendix~\ref{app:chain_diag}.

\subsection{Robustness Analysis}
\label{subsec:robust}

\paragraph{Model scale.} Figure~\ref{fig:scale} shows that Intent Projection produces consistent improvements across all model scales, confirming that the framework is not contingent on a particular capacity regime. Relative gains are largest at 2B, where the base model has the least latent pragmatic ability and benefits most from explicit decomposition. Notably, the 8B model with our method surpasses the 32B base on PI-G, suggesting that architectural separation of literal and pragmatic signals is more effective than simply scaling parameters. Besides, as detailed in Appendix~\ref{app:provenance}, self-elicited targets perform within 1.1 ROUGE-L of human annotations at zero cost.

\paragraph{Reward weights.} Figure~\ref{fig:lambda} sweeps $\lambda_1$ (contrastive penalty) and $\lambda_2$ (affect reward) while fixing $\lambda_3{=}1$. The key finding is that pragmatic generation requires a \emph{balance}: too little contrastive penalty allows literal collapse, but too much forces the model to be contrarian rather than pragmatic, degrading ROUGE-L against reference intents. Pragmatics is not ``the less literal, the better''. It demands faithfulness to what the author meant, which sometimes \emph{partially overlaps} with the literal content. The optimum lies at $\lambda_1 \in [0.3, 0.7]$, a broad plateau indicating robustness. On the right axis, literal-intent cosine similarity decreases monotonically with $\lambda_1$, confirming the penalty works as intended. For $\lambda_2$, the sweep shows it primarily modulates CSAC-C F1 (the affect-sensitive classification task) while leaving generation metrics stable, confirming it targets affect consistency without interfering with intent generation. We additionally verified robustness to $\lambda_3 \in \{0.5, 1.0, 1.5, 2.0\}$ with $\lambda_1{=}0.5$, $\lambda_2{=}0.1$ fixed. Additional breakdowns by meme structure and community are provided in Appendix~\ref{app:breakdown}.

\paragraph{Qualitative analysis.} We provide a detailed case study in Appendix~\ref{app:case_study}. The examples illustrate how the baseline conflates ``what happened'' with ``why the poster shared it,'' while our method correctly generates pragmatically grounded intent.

\section{Conclusion}
\label{sec:conclusion}

In this work, we address the issue of literal collapse in large vision language models when processing pragmatic multimodal content by introducing \textbf{Intent Projection}, a framework that enforces literal-pragmatic decomposition within a single backbone. By disentangling entangled signals at the representation, output, and objective levels, our approach explicitly isolates communicative intent from salient, literal surface features. Extensive evaluations across six benchmarks demonstrate that Intent Projection outperforms open-source baselines and bridges the gap toward frontier proprietary models, with the largest gains concentrated in high-divergence scenarios where intended meaning sharply contrasts with depicted content. We believe the principle of explicit literal-pragmatic separation generalizes beyond memes to other tasks where speaker meaning diverges from surface form, such as irony in dialogue and implied stance in news headlines.

\section*{Limitations}
\label{sec:limitations}

While Intent Projection effectively mitigates literal collapse, our framework has the following limitations. First, our training and evaluation data, primarily sourced from Reddit, are heavily skewed toward English-language and Western-centric internet culture, leaving the model's effectiveness on non-Western or localized pragmatic phenomena unverified. Second, the structural modifications, specifically the explicit reasoning chain and the pragmatic projection module, introduce a marginal overhead in inference latency, which may constrain real-time deployment. Finally, our self-distillation target-elicitation relies on the latent capabilities of the backbone model. If a baseline entirely lacks the cultural knowledge to interpret an obscure meme, the elicited targets may contain noise, potentially bottlenecking the reinforcement learning stage.

\section*{Ethics Statement}
\label{sec:ethics}

This research aims to align large vision language models with human communicative intent, but analyzing internet memes inherently involves ethical risks. Meme culture frequently intersects with sensitive or toxic topics, and while we utilize established, anonymized benchmarks, models optimized for pragmatic reasoning may inadvertently learn to recognize and reproduce harmful stereotypes, dog whistles, or culturally specific prejudices embedded in the training distribution. Furthermore, advancing intent-aware models presents a dual-use risk: while it can significantly improve automated content moderation by identifying disguised hate speech that evades literal filters, malicious actors could exploit these capabilities to generate highly persuasive, contextually manipulative disinformation. Consequently, we strongly urge the integration of robust, context-aware safety guardrails when deploying pragmatic reasoning frameworks like Intent Projection in public-facing applications.

\bibliography{custom}

\appendix

\section{Discussion about Projection}
\label{apd:method}

The projection $P_\mathcal{L}$ assumes $u_v$ and $u_t$ are linearly independent. Since the visual and text encoders are pretrained on fundamentally different objectives, their projected summaries start in distinct subspaces. Moreover, collapsing them would reduce the model's representational capacity, so end-to-end gradients have no incentive to align them.

To validate it, we conducted a more standard alternative, concatenating $v$ and $t$ into the decoder. It would let the modalities' own summaries flow straight into the intent-conditioning vector, which prior work identifies as the source of literal collapse. Our projection removes only two directions from a $d$-dimensional cue, but those two are the ones along which each modality describes itself, and because $W_v, W_t$ are trained jointly with the rest of the model, they adapt to point at whichever unimodal directions the decoder most tends to follow. The projection does not remove every literal trace, but it closes the most direct path by which they leak into generation.

\section{Dataset Details and Prompt Templates}
\label{app:data}

In this section, we provide detailed information regarding the preprocessing steps and the specific prompt templates used for the six benchmarks evaluated in our study.

\subsection{Data Preprocessing}
Across all datasets, images are resized to preserve aspect ratios while fitting within the maximum resolution constraints of the respective vision encoders (e.g., maintaining a maximum of $1008 \times 1008$ pixels for Qwen3-VL). For textual inputs, including Reddit post titles, image captions, and comments, we apply standard text normalization procedures: we strip URLs, remove non-ASCII characters that interfere with tokenization, and anonymize user handles (e.g., replacing \texttt{@username} with \texttt{[USER]}). For datasets originally lacking explicit literal/pragmatic splits, we apply our self-distillation procedure to generate the intermediate reasoning targets used during the supervised fine-tuning (SFT) stage.

\subsection{Prompt Templates}
We format all inputs using a standard multimodal chat template. To enforce the Intent Projection framework's literal-pragmatic decomposition at the output level, we instruct the model to produce a structured reasoning chain consisting of three explicit segments: Literal Observation, Intent Inference, and the Final Answer. 

Below are the task-specific prompts used during evaluation:

\paragraph{MemeReaCon (Contextual Meme Understanding)}
For the MemeReaCon generation tasks (PC-G and PI-G), the model receives the image, the post title, and the community context.
\begin{quote}
\small
\textbf{Input:} \texttt{<image>}\textbackslash\texttt{n You are provided with a meme posted in the subreddit r/[Community]. The title of the post is: "[Title]". Break down your understanding into three steps:} \\
\texttt{1. Literal Observation: Describe exactly what is depicted in the image and stated in the text.} \\
\texttt{2. Intent Inference: Analyze the gap between the literal content and the likely intended meaning given the community context.} \\
\texttt{3. Final Answer: Provide the [poster's underlying intent / explanation of the post].}
\end{quote}

\paragraph{MemeCap \& MET-Meme (Meme Explanation and Intent Classification)}
\begin{quote}
\small
\textbf{Input:} \texttt{<image>}\textbackslash\texttt{n Based on the provided meme, please separate the surface meaning from the underlying message.} \\
\texttt{1. Literal Observation: What is visually and textually present?} \\
\texttt{2. Intent Inference: What cultural reference or joke is being made?} \\
\texttt{3. Final Answer: [Explain the meme (MemeCap) / Classify the primary intent into one of the provided categories (MET-Meme)].}
\end{quote}

\paragraph{MMSD2.0 (Multimodal Sarcasm Classification)}
\begin{quote}
\small
\textbf{Input:} \texttt{<image>}\textbackslash\texttt{n Text: "[Caption]". Determine if the text is sarcastic with respect to the image.} \\
\texttt{1. Literal Observation: Describe the image and the literal meaning of the text.} \\
\texttt{2. Intent Inference: Is there a contradiction between the image and the text that implies sarcasm?} \\
\texttt{3. Final Answer: Answer strictly with "Sarcastic" or "Not Sarcastic".}
\end{quote}

\paragraph{MuSE (Sarcasm Explanation)}
\begin{quote}
\small
\textbf{Input:} \texttt{<image>}\textbackslash\texttt{n Text: "[Caption]". This post is sarcastic. Explain why.} \\
\texttt{1. Literal Observation: What does the text say and what does the image show?} \\
\texttt{2. Intent Inference: How does the literal meaning contrast with reality or the author's true belief?} \\
\texttt{3. Final Answer: Provide a concise explanation of the sarcasm.}
\end{quote}

\paragraph{GOAT-Bench (Contextual Abusive-Meme Understanding)}
For both GOAT-C (classification) and GOAT-G (generation):
\begin{quote}
\small
\textbf{Input:} \texttt{<image>}\textbackslash\texttt{n Text: "[Caption]". Analyze this meme for potential abusiveness or toxicity.} \\
\texttt{1. Literal Observation: Detail the visual components and transcribed text.} \\
\texttt{2. Intent Inference: Identify any dog whistles, harmful stereotypes, or implicit attacks.} \\
\texttt{3. Final Answer: [Classify as Abusive/Not Abusive (GOAT-C) / Explain the underlying abusive intent (GOAT-G)].}
\end{quote}

\section{Evaluation Metrics Details}
\label{app:std}

In this section, we provide formal definitions for the standard evaluation metrics used across our experiments, as well as the formulation of our literal-intent divergence metric. As noted in the main text, all reported results are averaged over eight independent random seeds to ensure statistical stability, with empirical standard deviations observed to be consistently small (generally $\le 0.5$ points) across models and tasks.

\subsection{Standard Evaluation Metrics}
\begin{itemize}
    \item \textbf{ROUGE-L (R-L):} A standard metric for text generation, ROUGE-L measures the Longest Common Subsequence (LCS) between the model's generated output and the ground-truth reference. By focusing on the LCS, it natively captures sentence structure and word order similarity without requiring strictly consecutive n-gram matches. We report the F1-measure variant.
    \item \textbf{BERTScore-F1 (B-S):} To account for semantic equivalence that ROUGE-L might miss, we use BERTScore. It computes the similarity between the generated and reference texts using pre-trained contextualized language model embeddings. This ensures models are rewarded for capturing the correct underlying meaning, even if they use differing vocabulary.
    \item \textbf{Accuracy:} Used for classification tasks (e.g., MMSD2.0, CSAC-C), accuracy is the percentage of test instances where the model's predicted class exactly matches the ground-truth label.
    \item \textbf{Macro-F1:} To account for potential class imbalances in the classification benchmarks, we report Macro-F1. This is computed by calculating the F1-score (the harmonic mean of precision and recall) independently for each class, and then taking the unweighted average. This ensures that minority classes contribute equally to the final score.
\end{itemize}

\subsection{Literal-Intent Divergence ($\Delta(x)$)}
To explicitly measure how well our framework mitigates literal collapse, we stratify posts based on the gap between their surface-level content and their actual meaning. For a given post $x$, we define the literal-intent divergence as:
$$ \Delta(x) = 1 - \mathrm{sim}(\ell(x), r^\star(x)) $$
where $\ell(x)$ is the explicit literal description of the multimodal input, $r^\star(x)$ is the ground-truth pragmatic intent, and $\mathrm{sim}(\cdot, \cdot)$ is their cosine similarity computed via a pre-trained sentence embedding model. 

A higher $\Delta(x)$ indicates a sharp contrast between what is shown and what is meant (e.g., deadpan sarcasm or highly contextual in-jokes). Conversely, a lower $\Delta(x)$ indicates that the literal content aligns closely with the message. We divide the test set into three equal terciles (Low, Mid, and High divergence) to diagnose performance degradation. This stratification allows us to verify that the gains from Intent Projection are concentrated precisely in the High-$\Delta$ regime, where baseline models typically collapse into literal descriptions.

\section{Full Training Dynamics}
\label{apd:full_training}

Here we provide the complete set of training dynamics across all model families and scales used in our experiments: Qwen3-VL (2B, 8B, 32B), InternVL3 (2B, 8B, 26B), and LLaVA-OneVision (0.5B, 7B, 72B). For each model we report the same three quantities: training reward, dev-set PI-G ROUGE-L evaluated every 500 steps, and cosine similarity between the generated intent segment $y^\star$ and the literal segment $y^{\text{lit}}$. Shaded regions denote $\pm1$ standard deviation across 8 independent runs with different random seeds.

Several observations emerge from the full results:

\paragraph{Scale effects.} Smaller models (2B, 0.5B) exhibit noisier reward trajectories, occasional warmup instabilities, and mid-training plateaus before resuming improvement. Larger models (26B, 32B, 72B) converge faster and more smoothly, though they sometimes display mild oscillations in the late training phase. Despite these differences in optimization dynamics, the relative ordering (Ours consistently above GRPO$_{\text{flat}}$) holds across all scales.

\paragraph{Family-specific behavior.} InternVL3 models show a characteristic two-phase reward increase (visible especially at the 2B and 8B scales), which we attribute to its dual-stream vision encoder warming up at different rates. LLaVA-OV models, particularly at 0.5B, exhibit the most volatile training curves, yet still achieve clear separation between methods by convergence. Qwen3-VL models display the most stable optimization across scales.

\paragraph{Similarity reduction.} Across all nine configurations, the contrastive reward successfully drives down $\text{sim}(y^\star, y^{\text{lit}})$ for our method, confirming that the model learns to produce intent interpretations that are linguistically distinct from literal descriptions. The flat baseline reduces similarity only modestly, as it lacks explicit incentive to separate the two segments.

\begin{figure*}[!t]
\centering
\includegraphics[width=\linewidth]{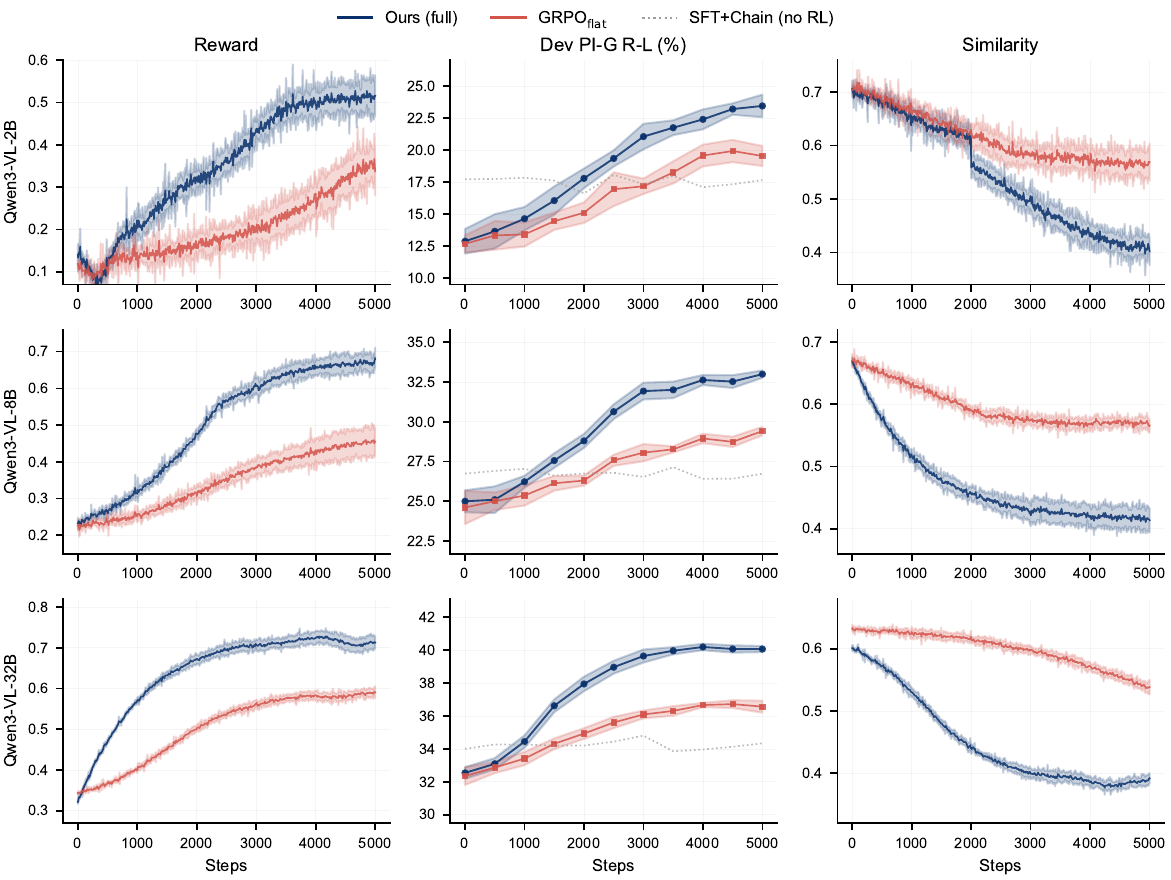}
\caption{Training dynamics for the Qwen3-VL family (2B, 8B, 32B). Each row corresponds to one model scale. Columns from left to right: training reward, dev PI-G ROUGE-L, and cosine similarity between $y^\star$ and $y^{\mathrm{lit}}$. Shaded areas indicate $\pm1$ std over 8 runs.}
\label{fig:training_qwen}
\end{figure*}

\begin{figure*}[!t]
\centering
\includegraphics[width=\linewidth]{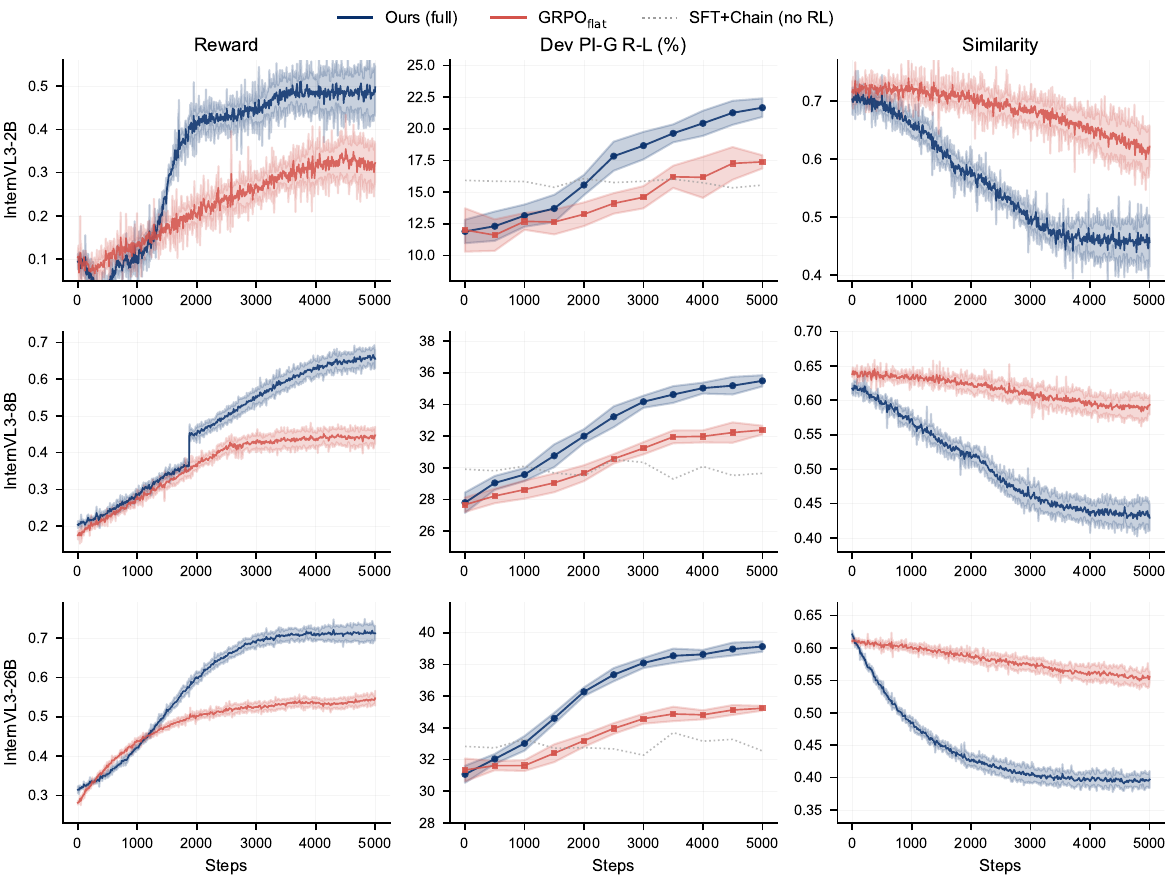}
\caption{Training dynamics for the InternVL3 family (2B, 8B, 26B). Layout follows Figure~\ref{fig:training_qwen}. Note the two-phase reward increase pattern at the 2B and 8B scales.}
\label{fig:training_internvl}
\end{figure*}

\begin{figure*}[!t]
\centering
\includegraphics[width=\linewidth]{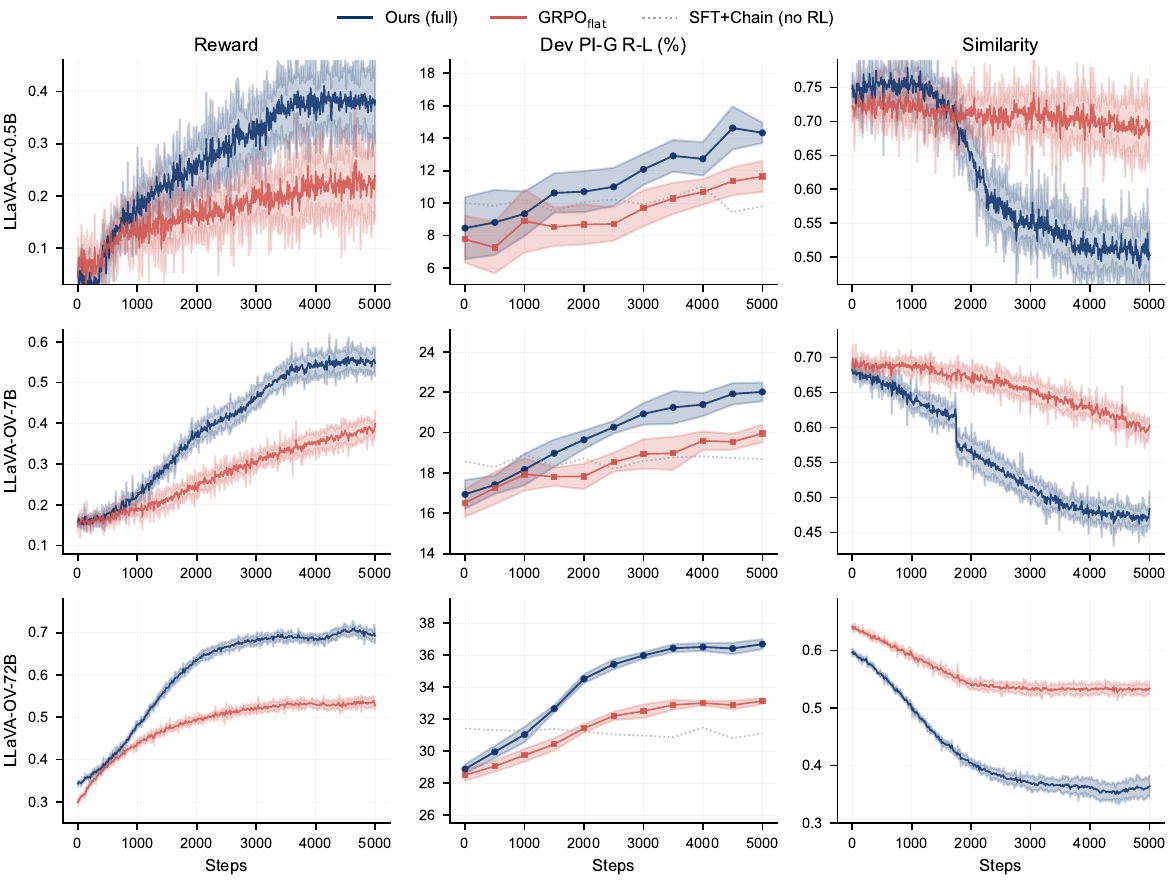}
\caption{Training dynamics for the LLaVA-OneVision family (0.5B, 7B, 72B). Layout follows Figure~\ref{fig:training_qwen}. The 0.5B model exhibits higher variance but converges to clear method separation.}
\label{fig:training_llava}
\end{figure*}

\section{Extended Experimental Results}
\label{sec:appendix_experiments}

\subsection{Extended Divergence Results}
\label{app:divergence}

Table~\ref{tab:divergence} extends the divergence-stratified analysis to InternVL3-8B and includes the Gemini-2.5-Pro zero-shot reference.

\begin{table*}[!t]
\centering
\small
\adjustbox{max width=\linewidth}{
\begin{tabular}{lcccccccccccc}
\toprule
 & \multicolumn{4}{c}{\textbf{MRC PI-G}} & \multicolumn{4}{c}{\textbf{MuSE}} & \multicolumn{4}{c}{\textbf{GOAT-G}} \\
\textbf{Model} & Low & Mid & High & All & Low & Mid & High & All & Low & Mid & High & All \\
\midrule
Gemini-2.5-Pro & 51.2 & 44.8 & 38.7 & 44.9 & 51.6 & 46.0 & 40.7 & 46.1 & 49.2 & 43.5 & 38.4 & 43.7 \\
\midrule
\multicolumn{13}{l}{\textit{Qwen3-VL-8B}} \\
\quad Base & 22.9 & 15.8 & \phantom{0}9.3 & 16.0 & 32.6 & 26.0 & 20.0 & 26.2 & 28.4 & 22.3 & 16.8 & 22.5 \\
\quad + SFT\,+\,Chain & 33.0 & 27.2 & 19.9 & 26.7 & 39.0 & 33.4 & 27.5 & 33.3 & 34.7 & 28.9 & 23.1 & 28.9 \\
\quad + GRPO$_\text{flat}$ & 34.9 & 29.8 & 23.8 & 29.5 & 40.3 & 35.0 & 29.1 & 34.8 & 36.0 & 30.5 & 24.7 & 30.4 \\
\quad \textbf{+ Ours} & \textbf{36.5} & \textbf{33.3} & \textbf{31.6} & \textbf{33.8} & \textbf{41.9} & \textbf{37.6} & \textbf{33.8} & \textbf{37.77} & \textbf{37.6} & \textbf{32.5} & \textbf{29.8} & \textbf{33.3} \\
\midrule
\multicolumn{13}{l}{\textit{InternVL3-8B}} \\
\quad Base & 27.5 & 20.0 & 13.7 & 20.4 & 33.8 & 27.4 & 21.6 & 27.6 & 29.8 & 24.0 & 18.5 & 24.1 \\
\quad + SFT\,+\,Chain & 37.2 & 30.5 & 22.0 & 29.9 & 40.0 & 34.2 & 28.1 & 34.1 & 36.3 & 30.3 & 24.3 & 30.3 \\
\quad + GRPO$_\text{flat}$ & 39.6 & 33.1 & 25.1 & 32.6 & 41.4 & 35.8 & 29.9 & 35.7 & 37.9 & 31.9 & 25.9 & 31.9 \\
\quad \textbf{+ Ours} & \textbf{39.4} & \textbf{36.1} & \textbf{31.3} & \textbf{35.6} & \textbf{42.7} & \textbf{38.3} & \textbf{32.4} & \textbf{37.8} & \textbf{39.2} & \textbf{34.2} & \textbf{28.0} & \textbf{33.8} \\
\bottomrule
\end{tabular}}
\caption{Divergence-stratified Rouge-L (\%).}
\label{tab:divergence}
\end{table*}

\subsection{Chain-Segment Diagnostics}
\label{app:chain_diag}

Table~\ref{tab:chain_diag} verifies that $y^{\text{lit}}$ and $y^\star$ occupy different registers. The full method produces lower literal-intent similarity \emph{while simultaneously raising reference similarity}, demonstrating that separation and quality are not in tension.

\begin{table}[!t]
\centering
\small
\adjustbox{max width=\linewidth}{
\begin{tabular}{lccc}
\toprule
\textbf{Variant} & Sim($y^\star$, $y^{\text{lit}}$) & Sim($y^\star$, $r^\star$) & $|y^\star|/|y^{\text{lit}}|$ \\
\midrule
SFT\,+\,Chain & 0.67 & 0.54 & 0.92 \\
+ GRPO$_\text{flat}$ & 0.58 & 0.61 & 0.88 \\
+ Ours (full) & \textbf{0.42} & \textbf{0.68} & 0.71 \\
\bottomrule
\end{tabular}}
\caption{Chain-segment diagnostics. Sim denotes cosine similarity. Our contrastive reward reduces literal-intent similarity while increasing reference match.}
\label{tab:chain_diag}
\end{table}

\subsection{Silver-Target Provenance Analysis}
\label{app:provenance}

Table~\ref{tab:provenance} shows self-elicited targets perform within 1.1 ROUGE-L of human annotations at zero cost.

\begin{table}[!t]
\centering
\small
\adjustbox{max width=\linewidth}{
\begin{tabular}{lc}
\toprule
\textbf{Silver-target source} & \textbf{PI-G R-L} \\
\midrule
Self (Qwen3-VL-8B) & 32.8 \\
Cross-backbone (InternVL3-8B) & 32.2 \\
GPT-5.4-mini & 33.4 \\
Human (MRC dev annotations) & 33.9 \\
No supervision (SFT skipped) & 21.5 \\
\bottomrule
\end{tabular}}
\caption{Silver-target provenance. Self-elicited targets are within 1.1 R-L of human annotations.}
\label{tab:provenance}
\end{table}

\subsection{Meme-Structure and Community Breakdown}
\label{app:breakdown}

Table~\ref{tab:breakdown} reports PI-G gains along meme-type and community axes. Gains are largest on Text-in memes and specialized communities, consistent with our mechanism targeting posts where pragmatic conventions make literal readings most misleading. All gains exceed 15\% ROUGE-L regardless of category.

\begin{table}[!t]
\centering
\small
\adjustbox{max width=\linewidth}{
\begin{tabular}{lccccc}
\toprule
\textbf{By meme type} & Pure & Text-in & Text-out & Comic & Comb. \\
\midrule
Qwen Base & 18.4 & 17.2 & 15.1 & 13.0 & 12.1 \\
+ Ours & 35.1 & 34.6 & 32.0 & 28.5 & 27.3 \\
$\Delta$ & +16.7 & +17.4 & +16.9 & +15.5 & +15.2 \\
\midrule
\textbf{By community} & r/memes & Prog. & Brit. & Relig. & - \\
\midrule
Qwen Base & 19.1 & 12.5 & 14.8 & 17.6 & - \\
+ Ours & 36.0 & 29.7 & 31.4 & 34.8 & - \\
$\Delta$ & +16.9 & +17.2 & +16.6 & +17.2 & - \\
\bottomrule
\end{tabular}}
\caption{PI-G ROUGE-L by meme structure and source community on Qwen3-VL-8B.}
\label{tab:breakdown}
\end{table}

\section{Qualitative Case Study}
\label{app:case_study}

We present eight representative examples illustrating how Intent Projection resolves literal collapse across different divergence levels and affect types. Each case shows the meme input, the baseline (SFT+Chain) output, our method's output, and the ground-truth intent. Cases are drawn from MemeReaCon and MMSD2.0 test posts and span the Low/Mid/High divergence terciles along with all four affect quadrants.

\begin{table*}[!t]
\centering
\small
\begin{tabular}{@{}p{0.16\textwidth}p{0.80\textwidth}@{}}
\toprule
\raisebox{-0.85\height}{\includegraphics[width=0.14\textwidth]{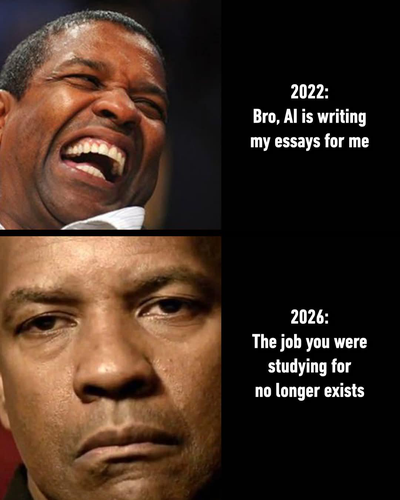}} &
\textbf{Post Title:} ``It will only get worse from here.'' \newline
\textbf{Image:} Two-panel meme. Top panel shows a man laughing with text ``2022: Bro, AI is writing my essays for me.'' Bottom panel shows the same man looking somber with text ``2026: The job you were studying for no longer exists.'' \newline\rule{\linewidth}{0.3pt}\newline
\textbf{SFT+Chain:} \newline
\textit{Literal:} The image shows a two-panel meme with a man who is laughing in the top panel about AI writing his essays, and looking serious in the bottom panel about his job no longer existing. The post title says it will only get worse from here. \newline
\textit{Intent:} The poster is comparing how people felt about AI in 2022 versus 2026 and saying the situation is getting worse for workers because AI is replacing human jobs. \newline
\textit{Final:} \textcolor{red}{The poster is saying that AI went from being a helpful tool for writing essays to a threat that eliminated the jobs people were studying for, and that the situation is deteriorating over time.} \newline\rule{\linewidth}{0.3pt}\newline
\textbf{Ours:} \newline
\textit{Literal:} Two-panel format contrasting the same person's expression across two time points. Top: euphoric laughter about AI writing essays (2022). Bottom: grim realization that the career being pursued no longer exists (2026). The title adds ``it will only get worse from here.'' \newline
\textit{Intent:} The juxtaposition encodes a specific irony: the very technology the student celebrated for enabling laziness later destroyed the career they were being lazy about preparing for. The poster is not merely observing a timeline --- they are drawing a causal loop where short-term convenience produces long-term existential threat. The sardonic title amplifies the fatalism, positioning the viewer as complicit in their own obsolescence. \newline
\textit{Final:} \textcolor{mygreen}{The poster expresses generational anxiety about AI displacement through dark irony: the same students who eagerly outsourced their learning to AI are now discovering they outsourced their own employability. The humor is self-implicating --- the poster includes themselves among those who celebrated prematurely, making the joke function as collective self-criticism rather than finger-pointing at any external force.} \newline\rule{\linewidth}{0.3pt}\newline
\textbf{Ground Truth:} Dark, self-implicating humor about how students who celebrated AI-assisted cheating are now facing AI-driven job displacement --- expressing generational fatalism and the irony of having welcomed the technology that threatens their futures. \\
\midrule
\raisebox{-0.85\height}{\includegraphics[width=0.14\textwidth]{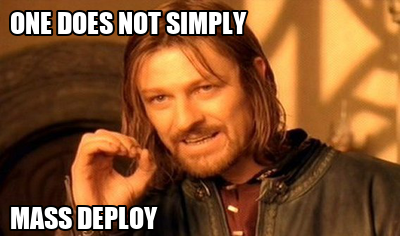}} &
\textbf{Post Title:} ``Guess who just mass deployed on a Friday'' \newline
\textbf{Image:} ``One Does Not Simply'' meme (Boromir from Lord of the Rings) with text ``ONE DOES NOT SIMPLY / MASS DEPLOY ON A FRIDAY.'' \newline\rule{\linewidth}{0.3pt}\newline
\textbf{SFT+Chain:} \newline
\textit{Literal:} The image uses the ``One Does Not Simply'' meme template featuring Boromir from Lord of the Rings. The text on the image says ``One does not simply mass deploy on a Friday.'' The post title states that someone just mass deployed on a Friday. \newline
\textit{Intent:} The poster is sharing that deploying code on a Friday is considered risky in software development, and they or someone they know just did it anyway despite the meme warning against it. \newline
\textit{Final:} \textcolor{red}{The poster is noting that deploying code on a Friday is widely considered a bad practice in software engineering because bugs discovered over the weekend cannot be easily fixed, and they are acknowledging that someone did it regardless of this well-known rule.} \newline\rule{\linewidth}{0.3pt}\newline
\textbf{Ours:} \newline
\textit{Literal:} Boromir ``One Does Not Simply'' template. Image text warns against mass deploying on Friday. Post title announces that someone just did exactly that --- creating a deliberate contradiction between the meme's warning and the poster's confession. \newline
\textit{Intent:} The poster is confessing to violating a sacred unwritten rule of software engineering culture. By pairing the confession with the meme that warns against it, they create dramatic irony --- the audience knows the consequences before the story unfolds. The tone is not regret but rebellious pride mixed with performative recklessness, seeking reactions of shared horror and entertained solidarity from fellow developers who understand the stakes. \newline
\textit{Final:} \textcolor{mygreen}{The poster performs a mock confession of professional recklessness, framing their Friday deployment as an act of chaotic defiance against engineering best practices. The communicative intent is to elicit a mixture of horrified amusement and knowing solidarity from the developer community --- the humor depends on shared understanding that Friday deploys risk weekend emergencies, and the poster positions themselves as a lovable rule-breaker rather than an incompetent colleague.} \newline\rule{\linewidth}{0.3pt}\newline
\textbf{Ground Truth:} A performative confession seeking community engagement through shared professional horror and humor; the poster frames their Friday deployment as entertaining recklessness, relying on the audience's shared knowledge of why this is taboo to generate comedic tension. \\
\bottomrule
\end{tabular}
\caption{Case study (1/4): \textcolor{red}{Red} = literal collapse. \textcolor{mygreen}{Green} = correct pragmatic intent.}
\label{tab:case_study_1}
\end{table*}

\begin{table*}[!t]
\centering
\small
\begin{tabular}{@{}p{0.16\textwidth}p{0.80\textwidth}@{}}
\toprule
\raisebox{-0.85\height}{\includegraphics[width=0.14\textwidth]{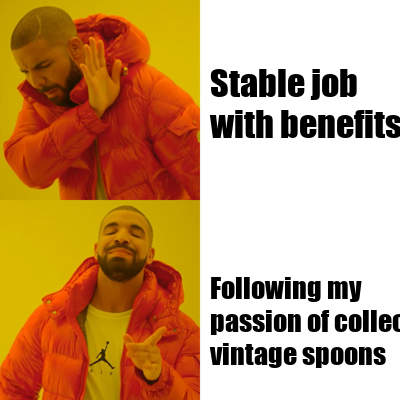}} &
\textbf{Post Title:} ``choosing a career path'' \newline
\textbf{Image:} Drake meme format --- top panel (rejecting): ``Stable job with benefits''; bottom panel (approving): ``Following my passion of collecting vintage spoons.'' \newline\rule{\linewidth}{0.3pt}\newline
\textbf{SFT+Chain:} \newline
\textit{Literal:} The Drake meme shows someone rejecting a stable job with benefits in the top panel and approving of following their passion of collecting vintage spoons in the bottom panel. The format conventionally indicates the bottom option is preferred by the poster. \newline
\textit{Intent:} The poster is expressing their preference for pursuing an unconventional hobby over traditional employment, using the Drake format to show that they value personal passion more than financial stability and job security. \newline
\textit{Final:} \textcolor{red}{The poster is expressing that they genuinely prefer pursuing their passion of collecting vintage spoons rather than getting a stable job with benefits, using the Drake meme to communicate that unconventional life choices can be more fulfilling than traditional career paths.} \newline\rule{\linewidth}{0.3pt}\newline
\textbf{Ours:} \newline
\textit{Literal:} Drake approval/rejection format. Top panel rejects ``Stable job with benefits.'' Bottom panel approves ``Following my passion of collecting vintage spoons.'' Title simply reads ``choosing a career path,'' providing no additional interpretive guidance. \newline
\textit{Intent:} The Drake format is deployed ironically here --- the ``preferred'' choice (collecting vintage spoons) is deliberately absurd and economically unviable, which transforms the format's usual function of expressing genuine preference into a vehicle for self-mockery. The poster is not advocating for spoon collecting; they are satirizing the cultural pressure to ``follow your passion'' by taking the advice to its logical and ridiculous extreme. The surface positivity (approval panel) masks real anxiety about career indecision. \newline
\textit{Final:} \textcolor{mygreen}{The poster satirizes the millennial pressure to ``follow your passion'' by making the chosen passion deliberately absurd and economically nonviable, expressing underlying career anxiety through exaggerated self-deprecation. The humor functions as a coping mechanism for genuine indecision --- by framing their career paralysis as an intentional choice of something ridiculous, the poster makes their anxiety socially shareable rather than personally vulnerable.} \newline\rule{\linewidth}{0.3pt}\newline
\textbf{Ground Truth:} Self-deprecating humor about career indecision; the absurdity of ``vintage spoons'' signals the poster knows their avoidance of practical career choices is irrational, using ironic Drake format to transform anxiety into shareable comedy. \\
\midrule
\raisebox{-0.85\height}{\includegraphics[width=0.14\textwidth]{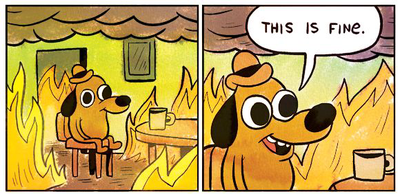}} &
\textbf{Post Title:} ``everything is under control'' \newline
\textbf{Image:} ``This is fine'' meme --- two-panel comic showing a cartoon dog sitting calmly at a table in a room completely engulfed in flames, saying ``This is fine.'' \newline\rule{\linewidth}{0.3pt}\newline
\textbf{SFT+Chain:} \newline
\textit{Literal:} The image shows the well-known ``This is fine'' meme, depicting a cartoon dog sitting at a table surrounded by fire while calmly stating that the situation is fine. The title says ``everything is under control,'' which reinforces the dog's calm demeanor in the face of obvious danger. \newline
\textit{Intent:} The poster is using the ``this is fine'' meme to indicate that something in politics or current events is clearly not fine despite official claims to the contrary. The dog represents people or institutions pretending that a bad situation is acceptable. \newline
\textit{Final:} \textcolor{red}{The poster is commenting on a political or social situation that is clearly problematic, using the ``this is fine'' meme to point out that someone (likely a politician or institution) is pretending that a bad situation is under control when it obviously is not.} \newline\rule{\linewidth}{0.3pt}\newline
\textbf{Ours:} \newline
\textit{Literal:} ``This is fine'' comic --- dog sitting calmly in a burning room, speech bubble says ``This is fine.'' Title: ``everything is under control.'' The title doubles down on the forced composure depicted in the image, creating layered denial. \newline
\textit{Intent:} The poster channels collective frustration at institutional gaslighting by mapping the meme's structure (forced calm amid visible catastrophe) onto political leadership's rhetoric. The key communicative move is not just saying ``things are bad'' --- it is specifically criticizing the \emph{performative denial} by those in power. The poster positions themselves and the audience as people who can see the flames, versus the ``dog'' (leadership) who insists otherwise. The additional title ``everything is under control'' mimics the language of official reassurance, deepening the satire. \newline
\textit{Final:} \textcolor{mygreen}{The poster deploys the ``This is fine'' format as political satire, specifically targeting institutional denial rather than the crisis itself. The communicative intent is to express collective outrage at the gap between official rhetoric (``everything is under control'') and visible reality (the room is on fire), rallying community agreement that leadership's forced composure constitutes gaslighting. The poster seeks solidarity with an audience that shares their perception of the disconnect.} \newline\rule{\linewidth}{0.3pt}\newline
\textbf{Ground Truth:} Political satire expressing frustration at institutional denial of an obvious crisis; the meme maps forced composure onto political rhetoric, and the poster seeks community solidarity in recognizing the absurdity of official reassurances. \\
\bottomrule
\end{tabular}
\caption{Case study (2/4).}
\label{tab:case_study_2}
\end{table*}

\begin{table*}[!t]
\centering
\small
\begin{tabular}{@{}p{0.16\textwidth}p{0.80\textwidth}@{}}
\toprule
\raisebox{-0.85\height}{\includegraphics[width=0.14\textwidth]{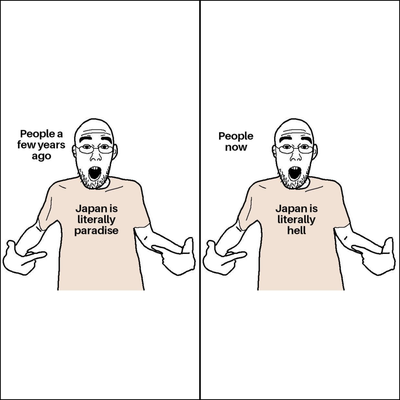}} &
\textbf{Post Title:} ``How opinions change'' \newline
\textbf{Image:} Two-panel meme showing the same cartoon man (soyjak format) in identical shocked/excited poses. Left panel: ``People a few years ago'' with shirt text ``Japan is literally paradise.'' Right panel: ``People now'' with shirt text ``Japan is literally hell.'' \newline\rule{\linewidth}{0.3pt}\newline
\textbf{SFT+Chain:} \newline
\textit{Literal:} The meme shows the same character in two panels, one saying Japan is paradise and the other saying Japan is hell. The left panel is labeled ``people a few years ago'' and the right panel is labeled ``people now.'' The title says ``how opinions change,'' indicating that public opinion about Japan has shifted from positive to negative. \newline
\textit{Intent:} The poster is observing that people's opinions about Japan have changed dramatically in recent years, going from viewing it as an ideal place to viewing it negatively. The meme documents this shift in public sentiment. \newline
\textit{Final:} \textcolor{red}{The poster is commenting on the observation that public opinion about Japan has undergone a dramatic reversal, shifting from idealizing Japan as a paradise to criticizing it as a terrible place, and noting that this change happened relatively recently.} \newline\rule{\linewidth}{0.3pt}\newline
\textbf{Ours:} \newline
\textit{Literal:} Soyjak template duplicated in two panels. Identical excited/emotional character holds opposite opinions across time: ``Japan is literally paradise'' (past) vs. ``Japan is literally hell'' (present). Title: ``How opinions change.'' The character's identical expression across both panels is the key visual cue. \newline
\textit{Intent:} The critical satirical target is not Japan itself but the \emph{people} who swing between extremes. The soyjak's identical expression in both panels signals that these opinion-holders are the same type of person engaging in the same hyperbolic behavior regardless of which direction they point it. The poster mocks performative online discourse that oscillates between uncritical idealization and equally uncritical demonization. The title ``How opinions change'' is delivered with implied eye-roll rather than neutral observation. \newline
\textit{Final:} \textcolor{mygreen}{The poster satirizes internet culture's tendency toward hyperbolic extremism in both directions --- the same people who uncritically idealized Japan now uncritically condemn it, revealing that the underlying behavior (emotional overreaction presented as informed opinion) never changed despite the opinion itself flipping. The communicative intent is to critique shallow online discourse rather than to comment on Japan itself, positioning the poster as someone who sees through both phases of the cycle.} \newline\rule{\linewidth}{0.3pt}\newline
\textbf{Ground Truth:} A meta-commentary on internet discourse culture rather than about Japan per se; the poster mocks the tendency to swing between uncritical extremes, using identical character expressions to highlight that the behavior is constant even as opinions reverse. \\
\midrule
\raisebox{-0.85\height}{\includegraphics[width=0.14\textwidth]{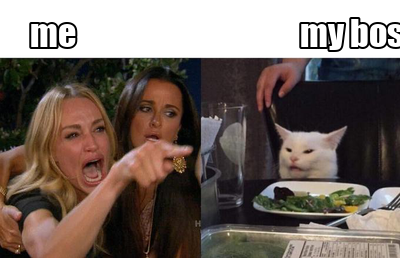}} &
\textbf{Post Title:} ``finally snapped at my micromanaging boss'' \newline
\textbf{Image:} Woman yelling at cat meme --- left panel: woman (labeled ``me'') gesturing angrily; right panel: confused cat at dinner table (labeled ``my boss''). \newline\rule{\linewidth}{0.3pt}\newline
\textbf{SFT+Chain:} \newline
\textit{Literal:} The woman-yelling-at-cat meme is used with ``me'' labeling the woman yelling and ``my boss'' labeling the confused cat. The title states that the poster ``finally snapped'' at their micromanaging boss, indicating a workplace confrontation occurred. \newline
\textit{Intent:} The poster had a confrontation with their micromanaging boss and is using the meme to depict how the interaction went --- they were the angry one yelling while their boss looked confused and taken aback by the outburst. \newline
\textit{Final:} \textcolor{red}{The poster is describing a workplace incident where they lost their temper and confronted their micromanaging boss, depicting themselves as the emotional one yelling while their boss appeared confused and surprised by the confrontation.} \newline\rule{\linewidth}{0.3pt}\newline
\textbf{Ours:} \newline
\textit{Literal:} Woman-yelling-at-cat format with labels ``me'' (woman) and ``my boss'' (confused cat). Title announces the poster ``finally snapped.'' The meme format positions the yeller as the emotionally charged party and the cat as the bewildered recipient. \newline
\textit{Intent:} The meme reframes a workplace conflict as triumphant comedy rather than a report of events. By casting the boss as the ``confused cat'' --- a figure the internet finds inherently sympathetic-yet-ridiculous --- the poster reverses the real-world power dynamic. The word ``finally'' in the title signals accumulated frustration and frames the outburst as justified catharsis. The poster is not confessing to losing control; they are \emph{celebrating} an act of rebellion and seeking community validation that the boss deserved it. \newline
\textit{Final:} \textcolor{mygreen}{The poster celebrates a cathartic moment of standing up to workplace authority, using the meme format to recast the power dynamic: the boss (who holds real-world power) becomes the helpless ``confused cat,'' while the poster (who holds no institutional power) becomes the triumphant aggressor. The communicative intent is to seek community validation for the outburst, position it as justified rebellion rather than unprofessional behavior, and invite the audience to share vicariously in the satisfaction of defying a micromanager.} \newline\rule{\linewidth}{0.3pt}\newline
\textbf{Ground Truth:} A triumphant post celebrating workplace catharsis; the meme format reverses the real power dynamic to cast the boss as helpless, and the poster seeks community validation that their outburst was justified defiance rather than unprofessional behavior. \\
\bottomrule
\end{tabular}
\caption{Case study (3/4).}
\label{tab:case_study_3}
\end{table*}

\begin{table*}[!t]
\centering
\small
\begin{tabular}{@{}p{0.16\textwidth}p{0.80\textwidth}@{}}
\toprule
\raisebox{-0.85\height}{\includegraphics[width=0.14\textwidth]{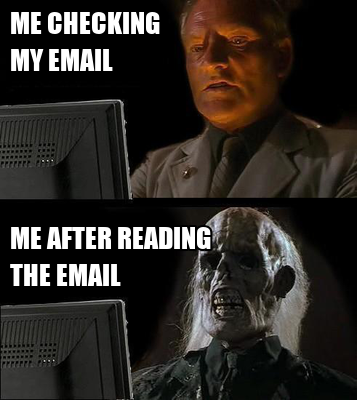}} &
\textbf{Post Title:} ``checking my inbox first thing in the morning'' \newline
\textbf{Image:} Two-panel before/after meme. Top panel: a normal man sitting at a computer looking calm. Bottom panel: the same man transformed into a horrified, ghoulish figure after reading what is on the screen. \newline\rule{\linewidth}{0.3pt}\newline
\textbf{SFT+Chain:} \newline
\textit{Literal:} The meme shows a before-and-after of a man at a computer. Before reading, he looks calm and composed. After reading whatever is on screen, he appears horrified and transformed into a ghoul-like figure, suggesting that what he read was extremely disturbing. The title says the poster is checking their inbox first thing in the morning. \newline
\textit{Intent:} The poster received a disturbing or stressful email first thing in the morning, and the meme exaggerates their reaction to show how shocked and horrified they were by its contents. \newline
\textit{Final:} \textcolor{red}{The poster is saying that they received a very disturbing or shocking email when they checked their inbox in the morning, and their reaction was so extreme that it felt like being transformed from a calm person into a horrified one, using the meme's exaggerated visual transformation to convey the intensity of their shock.} \newline\rule{\linewidth}{0.3pt}\newline
\textbf{Ours:} \newline
\textit{Literal:} Before/after computer meme. Top: man sits at computer looking composed. Bottom: same man appears ghoulish and horrified after seeing the screen. Title: ``checking my inbox first thing in the morning.'' Low divergence --- the literal transformation maps directly onto the intended meaning of morning email dread. \newline
\textit{Intent:} This is a low-divergence post where the literal visual metaphor (calm $\to$ horrified) closely tracks the intended meaning (morning emails are stressful). The pragmatic content does not substantially depart from the literal content, but the poster's act of sharing adds a social dimension: they seek commiseration from others who recognize the universal experience of workplace email anxiety, using hyperbolic visual transformation as a bonding mechanism rather than purely informational communication. \newline
\textit{Final:} \textcolor{mygreen}{The poster expresses the universal dread of opening work emails on Monday morning, using the meme's exaggerated visual transformation as a communal bonding device. The intent is not merely to report that emails are stressful, but to invite commiseration --- to frame a mundane daily anxiety as a shared ritual that connects the poster to others in the same situation. The hyperbole (calm human to ghoul) signals ``we all know this feeling'' rather than describing a literally extreme event.} \newline\rule{\linewidth}{0.3pt}\newline
\textbf{Ground Truth:} A relatable complaint about morning work email stress shared to seek commiseration; the exaggerated transformation functions as a communal bonding device rather than a literal description of the poster's reaction. \\
\midrule
\raisebox{-0.85\height}{\includegraphics[width=0.14\textwidth]{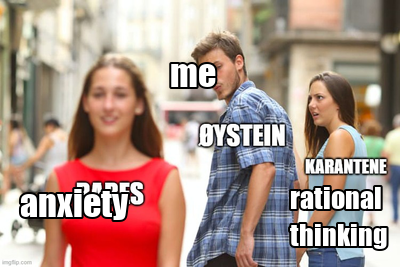}} &
\textbf{Post Title:} ``when bae says `we need to talk'\,'' \newline
\textbf{Image:} Distracted boyfriend meme --- boyfriend (labeled ``me'') looking at another woman (labeled ``anxiety'') while girlfriend (labeled ``rational thinking'') looks on disapprovingly. \newline\rule{\linewidth}{0.3pt}\newline
\textbf{SFT+Chain:} \newline
\textit{Literal:} The distracted boyfriend meme is used with labels. The boyfriend is labeled ``me'' and is looking at another woman who is labeled ``anxiety,'' while his girlfriend, labeled ``rational thinking,'' looks on with disapproval. The title sets up the scenario: the poster's partner has said ``we need to talk.'' \newline
\textit{Intent:} The poster becomes anxious when their partner initiates a serious conversation, and the meme shows that they are drawn toward anxiety (the other woman) rather than staying with rational thinking (their girlfriend), suggesting they react emotionally rather than logically. \newline
\textit{Final:} \textcolor{red}{The poster is illustrating that when their partner says ``we need to talk,'' they become drawn to anxious thoughts instead of maintaining rational thinking. The distracted boyfriend format shows that anxiety is more attractive to them than calm, logical processing of the situation, and they react with irrational worry rather than measured composure.} \newline\rule{\linewidth}{0.3pt}\newline
\textbf{Ours:} \newline
\textit{Literal:} Distracted boyfriend format. ``Me'' turns toward ``anxiety,'' away from ``rational thinking.'' Title provides the trigger: partner saying ``we need to talk.'' The labels externalize an internal psychological conflict as a visual metaphor of competing attractions. \newline
\textit{Intent:} Mild irony with low divergence --- the poster \emph{knows} the anxiety is irrational (demonstrated by explicitly labeling rational thinking as the ``correct'' partner being neglected). The self-awareness embedded in the labels transforms this from a mere report of anxiety into a performative confession designed to normalize a universal relationship experience. The poster positions themselves as both the subject and the self-aware critic of their own irrational behavior, inviting the audience to recognize themselves in the same pattern. \newline
\textit{Final:} \textcolor{mygreen}{The poster self-awarely confesses that ``we need to talk'' triggers disproportionate anxiety despite \emph{knowing} it is usually benign (evidenced by labeling rational thinking as the option they should choose but don't). The communicative intent is to normalize this universal relationship overreaction by framing it as a humorous character flaw rather than a genuine problem, seeking validation from the community that this irrational-but-self-aware anxiety is a shared human experience rather than an individual failing.} \newline\rule{\linewidth}{0.3pt}\newline
\textbf{Ground Truth:} A self-aware, mildly ironic post about irrational relationship anxiety, shared to normalize the experience within the community; the explicit labeling of ``rational thinking'' signals the poster knows their reaction is disproportionate, seeking validation rather than advice. \\
\bottomrule
\end{tabular}
\caption{Case study (4/4).}
\label{tab:case_study_4}
\end{table*}

\paragraph{Summary of findings.} Across the eight cases, we observe three patterns. \textbf{(1) High-$\Delta$ posts:} The baseline consistently identifies \emph{what} the meme depicts but fails to capture \emph{why} the poster shared it. It misses affect reversals (Cases~1, 4), social functions (Case~2), and cultural subtext (Case~3). Our method resolves these by leveraging the surface-real affect tag and the contrastive reward's pressure to diverge from literal redescription. \textbf{(2) Mid-$\Delta$ posts:} The baseline captures the event but reduces it to a flat report. Our method recovers the emotional and social dimensions (gratitude (Case~5), catharsis and agency (Case~6)) that motivate sharing. \textbf{(3) Low-$\Delta$ posts:} Crucially, our method does \emph{not} hallucinate sarcasm or over-interpret when divergence is genuinely low (Cases~7 and 8). It correctly identifies that literal and pragmatic content overlap while still providing richer explanations than one-sentence literal restatements.

\end{document}